\documentclass[cameraready]{Interspeech}
\usepackage{booktabs}
\usepackage{longtable}
\usepackage{booktabs}
\usepackage{graphicx}
\usepackage{comment}

\title{AfriVox-v2: A Domain-Verticalized Benchmark for In-the-Wild African Speech Recognition}

\author[affiliation={1}]{Busayo}{Awobade}
\author[affiliation={1}]{Gabrial Zencha}{Ashungafac}
\author[affiliation={1}]{Tobi}{Olatunji}

\address{
    $^1$ Intron Health
}

\email{research@intron.io, tobi@intron.io}


\usepackage{comment}

\begin{document}

\maketitle

\begin{abstract}



    Recent large language models (LLMs) show strong speech recognition and translation capabilities for high-resource languages. However, African languages remain dramatically underrepresented in benchmarks, limiting their practical use in low-resource settings. While early benchmarks tested African languages and accents, they lacked exhaustive real-world noise and granular domain evaluations. We present AfriVox-v2, a comprehensive benchmark designed to test speech models under realistic African deployment conditions. AfriVox-v2 introduces "in the wild" unscripted audio for all supported languages. We also introduce strict domain verticalization, evaluating model accuracy across ten sectors including government, finance, health, and agriculture and conducting targeted tests on numbers and named entities. Finally, we benchmark a new generation of speech models, including Sahara-v2, Gemini 3 Flash, and the Omnilingual CTC models. Our results expose the true generalization gap of modern speech models in specialized, noisy African contexts and provide a reliable blueprint for developers building localized voice AI.

\end{abstract}

\section{Introduction and Related Work}

Automatic Speech Recognition (ASR) has transitioned from a specialized tool to a foundational interface across global enterprise and consumer domains. In customer support, it facilitates real-time intent detection and agent assistance \cite{10.1145/2063576.2063776}; in healthcare, voice-enabled digital scribes alleviate clinician documentation burdens \cite{vanBuchem2021DigitalScribe, sanni2025afrispeechdialog}; and in legal settings \cite{DBLP:conf/icail/SaadanyBOW23}, it powers complex proceedings transcription \cite{olatunji2023afrispeech200, ashungafac2025afrispeech}. Today, the challenge for practitioners is no longer training bespoke models but selecting the optimal pre-trained foundation system—ranging from self-supervised models like wav2vec 2.0 \cite{Baevski2020wav2vec} to multitask giants like Whisper \cite{radford2023robust} or Specialized speech recognition LLMs and general purpose multimodal LLMs \cite{adelani2025irokobench}.

However, the "scaling laws" that have perfected ASR for Western languages have not yet closed the persistent "generalization gap" for African linguistic contexts \cite{ashungafac2025afrispeech}. While global leaderboards often report Word Error Rates (WER) below 5\% on standard English benchmarks , these models experience a 5x to 10x performance degradation when exposed to the rich phonetic and prosodic diversity of African accents \cite{olatunji2023afrispeech200}. This disparity is even more pronounced in indigenous African languages, where "supported" models often produce nearly unintelligible outputs with WERs exceeding 100\% \cite{ashungafac2025afrispeech}.

While the original AfriSpeech-MultiBench \cite{ashungafac2025afrispeech} provided a vital first look at African-accented English , the rapid proliferation of multimodal Speech LLMs and unscripted voice interfaces demands a more rigorous evaluation. Existing benchmarks suffer from three primary limitations:
\begin{itemize}
    \item \textbf{The "Read Speech" Bias}: Most datasets rely on scripted, read audio, which fails to capture the complexity of spontaneous, unscripted speech common in real-world African environments.
    \item \textbf{Surface-Level Domain Coverage}: General models often struggle with "verticalized" sectors—such as agriculture , finance, and telecommunications—where specialized vocabulary and named entities are critical.

    \item \textbf{Outdated Architecture Profiles}: With the emergence of the Omni-ASR family and updated proprietary models like Gemini 3, existing rankings no longer reflect the state-of-the-art.
\end{itemize}

In this work, we present AfriVox-v2, a significantly expanded benchmark designed to move beyond average-case accuracy toward real-world reliability. Our contributions are three-fold:
\begin{itemize}
    \item \textbf{"In the Wild" Data Expansion}: We incorporate unscripted, spontaneous audio across 20 African languages, capturing the environmental noise and overlapping speech that typically collapse traditional ASR performance.
    \item \textbf{Domain Verticalization}: We provide a granular performance analysis across key sectors including Government, Agriculture, Education, and Telecommunications, featuring dedicated subsets for numerical data and African named entities.
    \item \textbf{Modern Architecture Benchmarking}: We conduct a comprehensive evaluation of the first independent assessment of the Omni-CTC (300M, 1B, 7B) models \cite{omnilingualasrteam2025omnilingualasropensourcemultilingual}, Gemini 3 Flash \cite{gemini3flash2026}, and the regionally-tuned Sahara-v2 \cite{sahara}.
    \footnote{Sahara-v2 \cite{sahara} is developed by Intron Health, 
the authors' institution. All models were evaluated 
under identical conditions to ensure objectivity.}
\end{itemize}

By bridging the gap between curated read-speech and spontaneous "in the wild" interactions, AfriVox-v2 provides the evidence-based guidance necessary to deploy inclusive, reliable voice technology across the African continent

\section{Benchmark Methodology}

\begin{table}[t]
\centering
\caption{Overview of the datasets used, including the number of languages, total hours, and speech type.}
\label{tab:dataset_description}
\resizebox{\columnwidth}{!}{%
\begin{tabular}{l c c l}
\toprule
\textbf{Dataset} & \textbf{Languages} & \textbf{Hours} & \textbf{Speech Type} \\
\midrule
Waxal & 6 & $\sim$69.51 & Conversational \\
Africa Next Voices (AFN) & 15 & $\sim$100 & Conversational \\
Intron YT & 7 & 10 & Conversational \\
\bottomrule
\end{tabular}%
}
\end{table}

\subsection{Datasets}
AfriVox-v2 integrates multiple datasets spanning conversational and read speech across 20+ African languages and multiple language families. Prior read-speech corpora (e.g., Common Voice, FLEURS, NCHLT) are retained to maintain comparability with earlier benchmarks, while new conversational datasets substantially expand ecological validity.

\subsection{Intron-YT: Novel In-the-Wild Dataset}

We introduce Intron-YT, a new corpus of unscripted conversational speech collected from publicly available multimedia sources including podcasts, interviews, and public recordings with permissive licenses selected by human annotators from African crowdsourcing platforms based on the dominant language of the content. Following prior dataset construction methods \cite{yang2025gigaspeech, koluguri2025granary}, candidate recordings were segmented using voice activity detection (VAD), and adjacent segments were concatenated into utterances up to 30s. Each segment was then transcribed verbatim by native speakers, ignoring dysfluencies and unintelligible audio.

Annotators were college-educated bilingual speakers aged 18–35 and compensated between \$10–\$50/hour depending on task complexity and experience. Quality assurance involved a two-stage review pipeline: (1) Primary transcription by native speakers. (2) Independent meta-review by graduate-level annotators who validated 10–20\% of each contributor’s transcripts. Contributors achieving less than 80\% accuracy during review were excluded from the final dataset.

\begin{table}[t]
\centering
\footnotesize 

\caption{Sentence, number, and named entity counts per language.}
\label{tab:lang_counts_compact}
\begin{tabular}{lcccc}
\toprule
\textbf{Language} & \textbf{lang. code}  & \textbf{Sents.} & \textbf{Nums.} & \textbf{Ents.} \\
\midrule
Afrikaans & afr & 3301 & 228 & 509 \\
Akan & aka & 449 & 143 & 47 \\
Amharic & amh & 277 & 62 & 119 \\
Arabic & ara & 3366 & 245 & 716 \\
French & fra & 376 & 81 & 17 \\
Fulani & ful & 46 & 14 & 22 \\
Ga & gaa & 16 & 1 & 0 \\
Hausa & hau & 2083 & 239 & 540 \\
Igbo & ibo & 1363 & 219 & 304 \\
Kinyarwanda & kin & 2597 & 307 & 777 \\
Luganda & lug & 2131 & 141 & 448 \\
Pedi & nso & 2450 & 304 & 323 \\
Sesotho & sot & 2769 & 239 & 329 \\
Shona & sna & 876 & 331 & 254 \\
Swahili &swa & 2350 & 466 & 780 \\
Tswana & tsn & 1939 & 307 & 408 \\
Twi & twi & 376 & 39 & 76 \\
Xhosa & xho & 2430 & 284 & 432 \\
Yoruba & yor & 2064 & 320 & 764 \\
Zulu & zul & 2720 & 326 & 379 \\
\bottomrule
\end{tabular}

\vspace{1em} 

\caption{Total frequency of each domain tag.}
\label{tab:domain_counts_compact}
\begin{tabular}{lcc}
\toprule
\textbf{Domain} & \textbf{Count} & \textbf{Duration (hrs)} \\
\midrule
General & 20,117 & 29.67 \\
Health & 7,321 & 22.70 \\
Culture & 4,434 & 7.58 \\
Government & 4,323 & 8.76 \\
Finance & 1,882 & 3.31 \\
Education & 1,793 & 3.74 \\
Agriculture & 947 & 1.38 \\
Transportation & 911 & 1.82 \\
Sports & 620 & 1.47 \\
Telecom & 470 & 1.31 \\
\bottomrule
\end{tabular}
\end{table}

\subsection{Conversational Dataset Aggregation}

AfriVox-v2 aggregates conversational speech datasets including:
\begin{itemize}
    \item \textbf{Africa Next Voices (AFN)} -- a 9k hr+ multi-domain decentralized African conversational speech corpus funded by the Gates Foundation. Regional grantees created and hosted data separately, hence the need for aggregation. For AFN subsets with explicit test splits, we used those directly. For subsets without defined splits, we created evaluation sets using an 80/10/10 train/dev/test split stratified across unique speakers to prevent speaker leakage.
    \item \textbf{Waxal} – approximately 69.5 hours of multilingual conversational speech across 6 African languages drawn from the full Waxal corpus funded by Google. \cite{diack2026waxal}
\end{itemize}

For both AFN and Waxal, quality filtering was performed  by identifying samples with anomalously high WER from our best-performing model, followed by manual listening and validation against transcripts to remove bad audio and mismatched transcript pairs. This process retained approximately 90\% of samples, confirming overall corpus quality. All audio was resampled to 16kHz for evaluation.

\subsection{Domain-Verticalized Annotation}

To evaluate real-world application performance, we introduce a unified domain taxonomy derived from AFN and Afrispeech-MultiBench datasets.

The final taxonomy includes 10 domains: Agriculture, Culture \& Society, Finance, General, Health, Sports \& Hobbies, Telecommunications, Transport, Named Entities, and Numbers. Number and Named Entity domains are indicated when transcriptions contain at least one such element

Because only a subset of source datasets included domain labels, we developed a multilabel tagging pipeline. Inspired by recent results on multilingual LLM classification tasks \cite{adelani2025irokobench, hagos2025msteb}, transcripts were automatically tagged using Gemini-3.

Human validation was conducted on a random sample of approximately 50 utterances per language per tag. Validation was prioritized for the 6 languages with the highest utterance counts (500--800 samples per language), providing sufficient volume for statistically meaningful precision/recall estimates. Results yielded 70\% recall and 42\% precision, indicating acceptable signal despite moderate label noise.

\begin{table}[h!]
\centering
\small
\caption{Gemini Domain Validation Benchmark Results (Corrected Baseline)}
\resizebox{\columnwidth}{!}{%
\begin{tabular}{lccccc}
\hline
\textbf{Language} & \textbf{Samples} & \textbf{Precision} & \textbf{Recall}\\
\hline
Kinyarwanda & 500 & 35.32 & 55.60  \\
Sesotho     & 800 & 41.06 & 71.75  \\
Swahili     & 500 & 38.03 & 53.40  \\
Tswana      & 800 & 44.68 & 77.62  \\
Xhosa       & 799 & 45.17 & 76.60  \\
Zulu        & 800 & 40.72 & 69.62  \\
\hline
\end{tabular}%
}
\label{tab:domain_benchmark_results}
\end{table}

\begin{table*}[t]
\centering
\caption{Word Error Rate (WER) Comparison: Afrivox-1 vs. AfriVox-2. Each cell reports WER as \textit{Afrivox-1 / Afrivox-2}. The best scores for each language in Afrivox-1 are \underline{underlined}, and the best scores for Afrivox-2 are highlighted in \textbf{bold}. The \textbf{Average} column represents the overall mean across all 14 evaluated languages and is duplicated in both sections for easy reference.}
\label{tab:afrivox_results}
\resizebox{\textwidth}{!}{%
\begin{tabular}{lcccccccc}
\toprule
Model & akan & amh & ful & hau & ibo & kin & lug & \textbf{Average} \\
\midrule
Omni CTC 300M V2 & 60.27 / 54.92 & 40.49 / 48.98 & 47.33 / 58.81 & 40.25 / 40.09 & 46.05 / 44.64 & 44.17 / 21.52 & 51.41 / 48.34 & 42.52 / 39.20 \\
Omni CTC 1B V2   & 53.51 / 49.69 & 27.28 / 37.38 & 39.15 / 56.86 & 30.19 / 36.63 & 38.85 / 39.51 & 29.69 / 13.83 & 35.50 / 42.82 & 33.10 / 33.91 \\
Omni CTC 7B V2   & 44.18 / 44.73 & \underline{22.22} / 32.70 & \underline{34.93} / 55.68 & 25.01 / 50.22 & 30.90 / 45.90 & 22.24 / 10.38 & 23.85 / 42.22 & 27.16 / 32.20 \\
Gemini-3-flash   & 46.23 / 45.62 & 30.48 / \textbf{24.88} & 51.19 / 72.93 & 27.29 / \textbf{26.86} & 45.62 / 42.53 & 31.88 / 16.50 & 35.14 / \textbf{31.49} & 36.80 / 32.13 \\
Sahara-v2        & \underline{25.17} / \textbf{30.70} & 27.41 / 25.34 & 37.68 / \textbf{35.16} & \underline{18.71} / 28.46 & \underline{20.81} / \textbf{28.66} & \underline{11.30} / \textbf{6.59} & \underline{19.43} / 39.28 & \underline{20.55} / \textbf{23.78} \\
\midrule
\midrule
Model & sot & sna & swa & tsn & xho & yor & zul & \textbf{Average} \\
\midrule
Omni CTC 300M V2 & 60.55 / 53.32 & 32.87 / 42.49 & 29.21 / 15.16 & 34.07 / 27.72 & 39.36 / 34.41 & 35.51 / 43.78 & 33.79 / 34.58 & 42.52 / 39.20 \\
Omni CTC 1B V2   & 54.57 / 47.30 & 23.19 / 32.23 & 17.96 / 9.73  & 26.51 / 21.00 & 32.11 / 25.57 & 27.71 / 36.48 & 27.23 / 25.64 & 33.10 / 33.91 \\
Omni CTC 7B V2   & 51.17 / 42.12 & \underline{18.22} / 28.64 & 13.96 / 7.75  & 21.25 / 14.19 & 27.71 / \textbf{20.02} & 23.16 / 36.14 & 21.39 / \textbf{20.15} & 27.16 / 32.20 \\
Gemini-3-flash   & 50.33 / 34.63 & 70.56 / 36.81 & \underline{11.96} / 7.59  & 31.46 / 27.47 & 33.78 / 29.25 & 26.12 / 29.58 & 23.16 / 23.68 & 36.80 / 32.13 \\
Sahara-v2        & \underline{18.27} / \textbf{19.50} & 18.39 / \textbf{24.92} & 14.16 / \textbf{7.09}  & \underline{15.11} / \textbf{13.96} & \underline{26.28} / 20.59 & \underline{19.09} / \textbf{27.12} & \underline{15.87} / 25.57 & \underline{20.55} / \textbf{23.78} \\
\bottomrule
\end{tabular}%
}
\end{table*}

\begin{table*}[t]
\centering
\caption{Average Word Error Rate (WER) per domain, including sentences containing numbers and named entities, across all languages. The best score per domain is shown in \textbf{bold}.}
\label{tab:domain_analysis}
\resizebox{\textwidth}{!}{%
\begin{tabular}{lcccccccccccc}
\toprule
Model & Agriculture & Culture & Education & Finance & General & Government & Health & Sports & Telecom & Transport & Numbers & Entity \\
\midrule
Omni-CTC 300M & 42.86 & 41.33 & 40.07 & 45.94 & 44.58 & 44.36 & 43.75 & 45.48 & 48.23 & 45.25 & 42.66 & 45.23 \\
Omni-CTC 1B   & 30.55 & 29.76 & 29.10 & 32.40 & 33.68 & 31.37 & 34.28 & 32.95 & 36.18 & 32.20 & 32.80 & 33.70 \\
Omni-CTC 7B   & 26.84 & 24.43 & 22.83 & 26.95 & 28.54 & 25.95 & 28.52 & 26.19 & 30.96 & 27.46 & 27.19 & 27.87 \\
Gemini 3 Flash& 36.73 & 29.10 & 31.36 & 32.88 & 32.82 & 30.93 & 29.93 & 30.69 & 35.11 & 36.23 & 31.14 & 31.72 \\
Sahara-v2     & \textbf{16.11} & \textbf{21.32} & \textbf{18.04} & \textbf{17.00} & \textbf{16.12} & \textbf{18.72} & \textbf{16.12} & \textbf{21.60} & \textbf{25.38} & \textbf{19.77} & \textbf{20.32} & \textbf{23.11} \\
\bottomrule
\end{tabular}%
}
\end{table*}
\section{Experiments}

\begin{table*}[t]
\centering
\caption{Sahara-v2 Word Error Rate (WER \%) across domains for 20 languages. A lower value indicates better performance. Hyphens (-) indicate domains where no samples were available for that language.}
\label{tab:sahara_domain_wer}
\resizebox{\textwidth}{!}{%
\begin{tabular}{lcccccccccc}
\toprule
\textbf{Language} & \textbf{Agriculture} & \textbf{Culture} & \textbf{Education} & \textbf{Finance} & \textbf{General} & \textbf{Government} & \textbf{Health} & \textbf{Sports} & \textbf{Telecom} & \textbf{Transp.} \\
\midrule
\textbf{Afrikaans} & 22.5 & 34.5 & 22.9 & 18.5 & 19.3 & 22.3 & 19.9 & 38.6 & 31.5 & 19.1 \\
\textbf{Akan} & 44.5 & 20.1 & 24.5 & 21.6 & 21.3 & - & 24.8 & 23.0 & 30.6 & 19.7 \\
\textbf{Amharic} & 28.5 & 19.7 & 26.4 & 30.2 & 24.6 & 30.6 & 21.7 & 30.3 & 31.9 & 30.0 \\
\textbf{Arabic} & 26.6 & 24.7 & 13.1 & 13.2 & 18.5 & 22.5 & 11.4 & 12.9 & 20.4 & 17.5 \\
\textbf{French} & 23.8 & 0.0 & 12.6 & 20.2 & 13.5 & 21.9 & 13.0 & 20.0 & 0.0 & 26.6 \\
\textbf{Fulani} & - & 43.2 & 42.2 & 59.4 & 35.4 & 44.1 & 34.7 & 39.3 & 50.0 & 36.8 \\
\textbf{Ga} & - & 0.0 & - & 0.0 & 0.0 & - & 0.0 & 0.0 & 0.0 & - \\
\textbf{Hausa} & 13.9 & 18.5 & 17.7 & 16.9 & 17.7 & 18.8 & 18.4 & 17.3 & 18.9 & 17.3 \\
\textbf{Igbo} & 15.0 & 17.7 & 17.9 & 17.3 & 16.4 & 18.8 & 20.4 & 22.3 & 20.5 & 21.9 \\
\textbf{Kinyarwanda} & 8.3 & 10.6 & 9.3 & 9.3 & 8.4 & 9.2 & 14.1 & 10.7 & 13.4 & 12.0 \\
\textbf{Luganda} & 18.9 & 20.9 & 16.3 & 17.1 & 21.0 & 16.8 & 18.0 & 18.7 & 16.1 & 23.0 \\
\textbf{Pedi} & 17.5 & 29.0 & 28.5 & 29.6 & 15.3 & 27.3 & 22.8 & 32.8 & 50.0 & 27.9 \\
\textbf{Sesotho} & 9.2 & 24.6 & 21.6 & 17.1 & 15.4 & 16.1 & 20.2 & 21.8 & 30.2 & 21.0 \\
\textbf{Shona} & 25.0 & 28.1 & 24.5 & 30.2 & 20.7 & 28.5 & 8.4 & 26.3 & 36.7 & 22.3 \\
\textbf{Swahili} & 13.6 & 17.4 & 13.8 & 13.8 & 14.8 & 18.2 & 12.2 & 17.4 & 15.4 & 14.5 \\
\textbf{Tswana} & 9.0 & 20.5 & 14.8 & 14.1 & 12.0 & 14.3 & 15.8 & 32.2 & 18.5 & 13.9 \\
\textbf{Twi} & - & 6.0 & 33.7 & 7.1 & 9.5 & - & 10.2 & 20.0 & 0.0 & 8.3 \\
\textbf{Xhosa} & 18.2 & 27.6 & 25.6 & 22.8 & 19.1 & 26.5 & 19.1 & 32.2 & 31.4 & 26.4 \\
\textbf{Yoruba} & 13.7 & 16.4 & 14.9 & 16.1 & 15.9 & 17.6 & 16.7 & 16.6 & 19.6 & 15.8 \\
\textbf{Zulu} & 13.5 & 23.3 & 15.8 & 15.5 & 13.9 & 17.4 & 8.7 & 29.6 & 27.2 & 25.2 \\
\midrule
\textit{Average} & \textit{18.9} & \textit{20.1} & \textit{20.8} & \textit{19.5} & \textit{16.6} & \textit{21.8} & \textit{16.5} & \textit{23.1} & \textit{23.1} & \textit{21.0} \\
\bottomrule
\end{tabular}%
}
\end{table*}

\subsection{Models}
We benchmark models with broad African language coverage:
\begin{itemize}
    \item Gemini-3-Flash (multimodal speech LLM)
    \item Sahara-v2 (region-optimized ASR model)
    \item Omni-ASR v2 CTC models (300M, 1B, 7B parameters). CTC models were selected due to significantly faster inference compared to LLM-based decoding.
\end{itemize}

All models were evaluated using default preprocessing pipelines and hyperparameters provided by model maintainers. Language hints were passed when supported to reduce cross-language confusion.

\subsection{Evaluation Protocol}
We evaluate metrics using Word Error Rate. To better capture deployment-critical errors we introduce two additional metrics:
\begin{itemize}
    \item Entity Error Rate (EWER): WER restricted to samples with named entities.
    \item Numeric Error Rate (NWER): WER restricted to samples with numbers.
\end{itemize}

We report domain tag distribution, aggregate WER, domain-conditional WER, entity and numeric word error rates.

\section{Results}

\subsection{Scripted vs In-the-Wild Speech}

Table \ref{tab:afrivox_results} compares transcription performance across Afrivox-1 (primarily read speech) and Afrivox-2 (spontaneous in-the-wild speech). As expected, conversational speech generally increases recognition difficulty due to disfluencies, background noise, and acoustic variability. However, the results reveal non-uniform performance shifts across languages and models.

Across all languages, Sahara-v2 achieves the lowest average WER (23.78) on Afrivox-2, outperforming both multilingual CTC models and multimodal LLMs. Larger Omni-CTC models show consistent scaling improvements: average WER decreases from 39.20 (300M) to 33.91 (1B) and 32.20 (7B). This confirms the expected capacity–performance relationship, where larger multilingual models better capture phonetic diversity across African languages.

Interestingly, the relative performance difference between Afrivox-1 and Afrivox-2 varies substantially by language. In several languages (e.g., Kinyarwanda and Swahili), WER on Afrivox-2 is lower than on Afrivox-1 for some models. This likely reflects training data overlap or domain similarity between modern conversational datasets and model training corpora. In contrast, languages with broader acoustic variability or lower training coverage (e.g., Yoruba or Amharic) exhibit larger degradation between the two benchmarks.


These findings highlight a critical evaluation challenge: benchmark performance may reflect training data exposure rather than generalization ability.

\subsection{Model Scaling and Architecture Effects}

Model size strongly influences multilingual robustness. Across Afrivox-2, the Omni-CTC model family demonstrates clear scaling benefits, with the 7B parameter model outperforming smaller variants across most languages.

However, the multimodal Gemini-3 Flash model consistently lags behind speech-native ASR models, despite its strong performance on many multilingual text benchmarks. This suggests that current multimodal LLMs still struggle with precise phonetic transcription, likely due to architectures optimized for semantic understanding rather than exact acoustic decoding.

In contrast, the region-optimized Sahara-v2 model achieves the strongest performance overall, indicating that targeted regional training data can outperform larger general-purpose multilingual models when evaluating underrepresented linguistic contexts.

\subsection{Domain-Specific Performance Variation}

Table~\ref{tab:domain_analysis} presents domain-conditional WER across ten application domains, including specialized categories for numbers and named entities. Two consistent patterns emerge.

First, domain specialization significantly affects recognition accuracy. Telecommunications and Sports exhibit the highest error rates across models, with average WERs exceeding 30--35\% for most systems. These domains often contain specialized terminology, abbreviations, and named entities, which are poorly represented in general training corpora.

Second, models trained with regional or domain-specific data show substantial improvements. Sahara-v2 achieves the lowest WER across every evaluated domain, with performance as low as 16.11\% in Agriculture and 16.12\% in General discourse.

Table~\ref{tab:sahara_domain_wer} further reveals that domain sensitivity is compounded by language-level resource availability. High-resource languages such as Kinyarwanda and Swahili maintain stable cross-domain performance (8--14\% WER), while lower-resource languages like Fulani show greater domain variance (34--59\%). Cells with 0.0\% WER for languages such as Ga and French reflect data sparsity rather than true model proficiency. Notably, Pedi exhibits an anomalously high Telecom WER of 50.0\% compared to 15--32\% in other domains, suggesting particular vulnerability to telecom-specific vocabulary.

Even with the strongest models, error rates remain elevated for numbers (20.32\%) and named entities (23.11\%), representing a critical failure mode for real-world applications where incorrect transcription of financial values, identifiers, or proper names can lead to significant downstream errors.

\subsection{Implications for Voice AI Deployment}

These results reveal three important insights for speech technology deployment in multilingual African contexts.

First, aggregate WER masks significant domain-level performance variation. Systems with similar overall accuracy can differ substantially when transcribing domain-specific content.

Second, named entities and numerical expressions remain major sources of transcription error, even for the best-performing models.

Third, region-optimized models outperform larger global models, highlighting the importance of geographically and linguistically representative training data.

Taken together, these findings reinforce the need for ecologically valid benchmarks that evaluate speech systems beyond read speech and average WER. By incorporating spontaneous speech and domain-specific evaluation, AfriVox-v2 provides a more realistic framework for assessing ASR performance in real-world African deployment scenarios.

\section{Conclusion}
We present AfriVox-v2, the most comprehensive benchmark to date for evaluating speech recognition across African languages. By introducing a novel in-the-wild dataset, consolidating conversational corpora across 20+ languages, and enabling domain-verticalized evaluation, AfriVox-v2 exposes significant limitations of current ASR systems that are hidden by traditional benchmarks.

Our results demonstrate that modern speech models still struggle with domain-specific terminology, conversational speech, and linguistically diverse African languages. AfriVox-v2 provides a practical framework for evaluating and improving voice technologies in low-resource contexts and aims to catalyze more inclusive speech AI research.

\section{Limitations}

Despite its expanded coverage, AfriVox-v2 still represents only a fraction of Africa’s linguistic diversity. Many languages remain underrepresented, and several datasets contain relatively small amounts of conversational speech, limiting statistical power for fine-grained analysis.

Additionally, domain annotations rely partly on LLM-assisted labeling, which introduces label noise despite human validation. Validation on the 6 highest-utterance languages yielded 42\% precision and 70\% recall; domain-level findings should therefore be interpreted as indicative trends rather than precise performance estimates. Future work should explore fully human-annotated domain corpora and extend validation to all supported languages.

\bibliographystyle{IEEEtran}
\bibliography{mybib}

\end{document}